\documentclass[letterpaper, 10 pt, conference]{ieeeconf}  

\IEEEoverridecommandlockouts                              

\usepackage{bm}
\usepackage{url}
\usepackage{graphicx}
\usepackage{lipsum}
\usepackage{pifont}
\usepackage{amsmath}
\usepackage{amssymb}
\usepackage{nccmath}
\usepackage{comment}
\usepackage{balance}
\usepackage{multirow}
\usepackage{textcomp}
\usepackage{multirow}
\usepackage{subcaption}
\usepackage{booktabs}
\usepackage{algorithm}
\usepackage{dblfloatfix}
\usepackage[table]{xcolor}
\usepackage[noend]{algpseudocode}

\title{\LARGE \bf
Exact Point Cloud Downsampling \\ for Fast and Accurate Global Trajectory Optimization
}

\author{Kenji Koide$^{1}$, Shuji Oishi$^{1}$, Masashi Yokozuka$^{1}$, and Atsuhiko Banno$^{1}$
\thanks{*This work was supported in part by JSPS KAKENHI Grant Number 23K16979 and a project commissioned by the New Energy and Industrial Technology Development Organization (NEDO).}
\thanks{$^{1}$All the authors are with the Department of Information Technology and Human Factors, the National Institute of Advanced Industrial Science and Technology, Tsukuba, Ibaraki, Japan, {\tt\small k.koide@aist.go.jp}}%
}

\begin{document}

\maketitle
\thispagestyle{empty}
\pagestyle{empty}

\setlength\floatsep{8pt}
\setlength\textfloatsep{10pt}

\begin{abstract}

This paper presents a point cloud downsampling algorithm for fast and accurate trajectory optimization based on global registration error minimization. The proposed algorithm selects a weighted subset of residuals of the input point cloud such that the subset yields exactly the same quadratic point cloud registration error function as that of the original point cloud at the evaluation point. This method accurately approximates the original registration error function with only a small subset of input points (29 residuals at a minimum). Experimental results using the KITTI dataset demonstrate that the proposed algorithm significantly reduces processing time (by 87\%) and memory consumption (by 99\%) for global registration error minimization while retaining accuracy. 

\end{abstract}

\section{Introduction}

Global trajectory optimization is a crucial step for localization and mappin systems. Because it is unavoidable that trajectory errors accumulate in online odometry estimation that performs real-time optimization using local observations, it is necessary to correct estimation drift by considering the global consistency of the map.

Global registration error minimization is one of the most accurate approaches to global trajectory optimization \cite{lu1997globally}. Unlike the conventional pose graph optimization that minimizes the errors in the pose space \cite{Grisetti2010}, global registration error minimization directly minimizes the multi-frame point cloud registration errors over the entire map. This approach avoids the Gaussian approximation of the relative pose constraint and enables accurate trajectory optimization by jointly aligning all frames in the map \cite{Koide2021}. However, it is known to be computationally expensive compared to pose graph optimization, as it requires a re-evaluation of registration error functions that involve residual computations for many points \cite{Reijgwart2020}. It also consumes a substantial amount of memory in order to remember the point correspondences between frames.

\begin{figure}[tb]
  \centering
  \includegraphics[width=\linewidth]{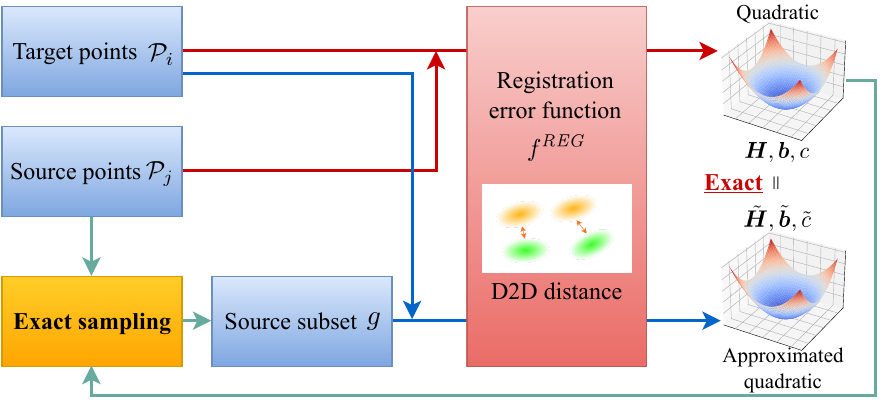}
  \caption{The proposed algorithm extracts a weighted subset of input source points such that the subset yields the same quadratic registration error function as that of the original points at the evaluation point.}
  \label{fig:teaser}
\end{figure}

To mitigate the processing cost and memory consumption of global registration error minimization, we propose a point cloud downsampling algorithm based on an efficient and exact weighted coreset extraction algorithm \cite{NEURIPS2019_475fbefa}. A coreset is a subset of an input dataset selected such that the result of an algorithm on the coreset approximates that on the original set \cite{phillips2017coresets}. For example, considering a Hessian matrix calculated from a Jacobian matrix (${\bm H} = {\bm J}^T {\bm J}$ and ${\bm J} \in \mathbb{R}^{N \times D}$), a weighted coreset ${\bm S} \in \mathbb{R}^{M \times D}$ approximates the original Hessian matrix with a subset of the Jacobian matrix (${\bm S}^T {\bm S} \approx {\bm H}$ and ${\bm S} \subset {\bm J}$). In this work, we employ a novel coreset extraction algorithm, which finds a coreset to {\it exactly} represent the original Hessian matrix in a time linear to the number of points \cite{NEURIPS2019_475fbefa}.

\begin{figure}[tb]
  \centering
  \includegraphics[width=1.0\linewidth]{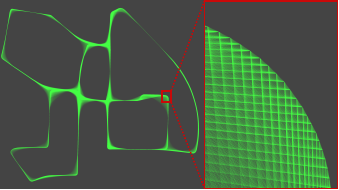}
  \caption{Dense factor graph for global registration error minimization. The proposed algorithm reduces memory consumption by 99\% and processing time by 87\% for the optimization of the factor graph.}
  \label{fig:graph}
\end{figure}

Using this algorithm, we find a coreset of the residuals of the input point cloud such that it exactly reconstructs the original quadratic registration error function at the evaluation point (see Fig. \ref{fig:teaser}). The proposed method needs only 29 residuals at a minimum to compose an exact coreset for a quadratic error function with six-dimensional input. It is also able to find a larger coreset with M residuals (e.g., M=256) to enhance the nonlinearity approximation accuracy. Note that because one three-dimensional point yields three residuals for point cloud registration, the computation of 29 residuals entails a processing cost approximately equal to that for 10 points.

For global trajectory optimization, we construct an extremely dense global registration error minimization factor graph, as shown in Fig. \ref{fig:graph}. The graph contains 4,540 pose variables and 585,417 registration error factors. We evaluate the residuals of approximately 10,000 points for each factor at every optimization iteration. The optimization requires approximately 22 GB of memory and 13 hours on a CPU with 128 threads. The proposed algorithm drastically decreases the number of residuals to be computed and reduces memory consumption and processing time to approximately 0.25 GB and 1.7 hours, respectively, while retaining estimation accuracy.

The main contributions of this work can be summarized as follows:
\begin{itemize}
  \item We extend the exact coreset extraction algorithm in \cite{NEURIPS2019_475fbefa} so that it can extract coresets with an arbitrary number of residuals. We also made several modifications to improve the processing speed of the algorithm.
  \item Based on the extended exact coreset extraction algorithm, we propose a point cloud downsampling algorithm that exactly recovers the original quadratic registration error function at the evaluation point. It also shows good approximation accuracy for the nonlinearity of the original function.
  \item We release the code of the proposed coreset extraction algorithm as open source.
\end{itemize}

\section{Related Work}

\subsection{Pose Graph Optimization}

Pose graph optimization is the gold standard for global trajectory optimization for range-based SLAM \cite{Grisetti2010}. It models the relative pose between frames estimated by scan matching as a Gaussian distribution and optimizes the sensor poses via maximum a posteriori estimation in the pose space. To model the Gaussian relative pose constraint, an explicit estimate of the representative relative pose value (i.e., mean) and its uncertainty (i.e., covariance) is required. However, scan matching becomes unreliable when two frames have only a small overlap, making it difficult to obtain an accurate relative pose estimate, which, in turn, makes it difficult to constrain the relative pose between small overlapping frames with pose graph optimization \cite{Koide2021}. The uncertainty estimation of a scan matching result is also difficult in practice \cite{Landry2019}, and many works use inappropriate covariance matrices to construct the relative pose constraints (e.g., using constant covariance matrices \cite{behley2018efficient} or a simple weighting scheme \cite{Shan2020}). These difficulties in the modeling of the relative pose constraint lead to inaccurate estimation results.

\subsection{Bundle Adjustment}

Bundle adjustment (BA) is another approach to global trajectory optimization that simultaneously optimizes sensor poses and environment parameters. For range-based SLAM, BA is often formulated as the simultaneous estimation of pose variables and plane or edge environment parameters \cite{Wisth2023}. Because the plane and edge parameters can be estimated from sensor poses and point coordinates, they can be eliminated from the estimation variables, turning the BA problem into a direct eigenvalue minimization of the accumulated points \cite{Liu2021}. While this approach ensures a consistent mapping result, it can suffer from high computation cost and a small convergence basin due to the non-least-squares error function. To mitigate these issues, it is necessary to combine range-based BA with hierarchical pose graph optimization \cite{Liu2023}.

\subsection{Global Registration Error Minimization}

Global registration error minimization is an approach to directly minimize multi-frame registration errors over the entire map \cite{lu1997globally}. Because it avoids the Gaussian approximation of pose graph optimization, it enables the construction of accurate relative pose constraints between point clouds with a very small overlap. Furthermore, it naturally propagates the per-point uncertainty of input data to the pose uncertainty via re-linearization of point cloud registration errors. While it shows a substantially better accuracy compared to pose graph optimization, it is computationally expensive because it re-evaluates point cloud registration errors between all point cloud pairs at every optimization iteration. 

To mitigate the computation cost of global registration error minimization, one may consider reducing the number of input points by, for example, random sampling \cite{Reijgwart2020}. However, this may change the shape of the objective function, depending on the selected points, and can negatively and substantially affect the optimization result. Although there is another approach to speeding up global registration error minimization with GPU-acceleration, the computation cost is still high, and it requires submap-level registration \cite{Koide2021}.

\subsection{Coreset Extraction}

Geometric data summarization is an essential tool for handling big data in computational geometry, and coresets are one of the most important classes of summarization methods. A coreset is a subset of an input dataset selected such that the result of an algorithm on it approximates the result on the original set \cite{phillips2017coresets}. When a coreset produces the same output of an algorithm as that of the original set, it is called exact. For example, Caratheodory's theorem states that every point in a convex hull of points in $\mathbb{R}^D$ can be represented by a weighted sum of at most $D + 1$ points \cite{cook1972caratheodory}. This means there always exists an exact coreset with only $D + 1$ points to represent a point in a convex hull (a.k.a., a Caratheodory set). 

While Caratheodory provided an algorithm to find such a coreset, the time required was $O(N^2D^2)$, rendering it impractical for large-scale problems \cite{cook1972caratheodory}. Although there were several subsequent works proposing efficient approximating algorithms to find a Caratheodory set, an exact algorithm with a linear time complexity had, until recently, been unavailable. However, in 2019, an epoch-making algorithm was proposed by Maalouf et al., one that finds an exact Caratheodory set in time linear to the number of input data points \cite{NEURIPS2019_475fbefa}. The method was applied to find an exact coreset to reconstruct a Hessian matrix using only a small subset of the input data for accelerating least squares optimization.

Our aim in this work is to introduce this novel data summarization technique to drastically reduce the computation cost of point cloud registration while retaining the accuracy of the original set by extracting an exact coreset of input points. To the best of our knowledge, this is the first work that introduces this type of exact data summarization technique in the context of point cloud registration.

\section{Methodology}

\subsection{Problem Setting}

Given a sequence of point clouds $\mathcal{P}_i = \{ {\bm p}_1, \cdots, {\bm p}_N \} $ and an initial guess of their poses $\breve {\bm T}_i$, we estimate refined sensor poses ${\bm T}_i$ by minimizing the registration errors between all overlapping point clouds. The registration error function $f^{\text{\it REG}}$ is defined as the sum of least square errors between corresponding points:

\begin{align}
f^{\text{\it REG}}(\mathcal{P}_i, \mathcal{P}_j, {\bm T}_i, {\bm T}_j) &= \sum_{{\bm p}_k \in \mathcal{P}_j} \| f^{\text{\it DIST}} ({\bm p}_k', {\bm p}_k, {\bm T}_i, {\bm T}_j) \|^2, \\
&= {\bm e}^T {\bm e},
\end{align}
where ${\bm p}_k' \in \mathcal{P}_i$ is the corresponding point of ${\bm p}_k \in \mathcal{P}_j$, $f^{\text{\it DIST}}$ is a distance function between points that returns a residual vector ${\bm e}_{k} \in \mathbb{R}^3$, and ${\bm e} = [{\bm e}_1^T, {\bm e}_2^T, \cdots]^T$ is a stack of ${\bm e}_{k}$. 

In Gauss-Newton optimization, the error function $f^{\text{\it REG}}$ is linearized at the current estimate $\breve {\bm T} = (\breve {\bm T}_i, \breve {\bm T}_j)$ and modeled in the quadratic form

\begin{align}
\label{eq:quadratic}
f^{\text{\it REG}}(\mathcal{P}_i, \mathcal{P}_j, \breve {\bm T} \boxplus \Delta {\bm x}) &\approx  \Delta {\bm x}^T {\bm H} \Delta {\bm x} + 2 {\bm b}^T \Delta {\bm x} + c,
\end{align}
where $c = f^{\text{\it REG}}(\mathcal{P}_i, \mathcal{P}_j, \breve {\bm T})$ is a constant, ${\bm J} = \partial {\bm e} / \partial {\bm T}$ is the Jacobian of residuals evaluated at $\breve {\bm T}$, and ${\bm b} = {\bm J}^T {\bm e}$ and ${\bm H} = {\bm J}^T {\bm J}$ are the information vector and matrix, respectively.

Because the linearization of $f^{\text{\it REG}}$ involves residual computation for all input points in $\mathcal{P}_j$, it is computationally expensive. It also consumes a substantial amount of memory to remember the corresponding points for each point cloud pair.

To mitigate the memory consumption and processing cost of global registration error minimization, we introduce a downsampling function that takes as input the error function, point clouds, and current estimate of sensor poses, and outputs a weighted subset of residuals:
\begin{align}
f^{\text{\it DOWN}}(f^{\text{\it REG}}, \mathcal{P}_i, \mathcal{P}_j, \breve {\bm T}_i, \breve {\bm T}_j) = ({\bm w}_{ij}, g_{ij}),
\end{align}
where $g_{ij}({\bm e}) = \tilde{{\bm e}}$ is a function to select $M$ residuals from ${\bm e}$ (i.e., $\tilde{\bm e} \subset {\bm e}$ and $\tilde{\bm e} \in \mathbb{R}^M$), and ${\bm w}_{ij} \in \mathbb{R}^M$ is the weights for the selected residuals. During optimization, we use the subset of residuals $\tilde{\bm e}$ instead of the original set to re-linearize the nonlinear error function $f^{\text{\it REG}}$.

To accurately approximate the original error function with only a small subset of the input data, we select a subset that will yield exactly the same quadratic error function as the original function at the evaluation point $\breve {\bm x}$:

\begin{align}
{\bm H} = \tilde{\bm H}, \qquad
{\bm b} = \tilde{\bm b}, \qquad
c = \tilde{c},
\end{align}
where
\begin{align}
\tilde{\bm e} = g_{ij} \left( {\bm e} \right), \quad
\tilde{\bm J} = \frac{\partial \tilde{\bm e}}{\partial {\bm T}}, \quad
{\bm W}_{ij}  = \text{diag}({\bm w}_{ij}), \\
\tilde{\bm H} = \tilde{\bm J}^T {\bm W}_{ij} \tilde{\bm J}, \quad
\tilde{\bm b} = \tilde{\bm J}^T {\bm W}_{ij} \tilde{\bm e}, \quad
\tilde{c}     = \tilde{\bm e}^T {\bm W}_{ij} \tilde{\bm e}.
\end{align}

For simplicity and efficiency, we assume that the initial guess of sensor poses is reasonably accurate and use the point correspondences found at the initial state during optimization.

\subsection{Registration Error Function}

The proposed downsampling algorithm only requires the error function to be a first-order differentiable least squares function. It thus can be applied to most of the common registration error functions (e.g., point-to-point \cite{Chetverikova} and point-to-plane \cite{Xu2022} distance functions). In this work, we use the generalized ICP (GICP) error function \cite{segal2009generalized} based on the distribution-to-distribution distance.

GICP models each point as a Gaussian distribution ${\bm p}_k = ({\bm \mu}_k, {\bm C}_k)$ representing the local surface shape around the point and calculates the distribution-to-distribution distance between a point ${\bm p}_k \in \mathcal{P}_j $ and its corresponding point ${\bm p}'_k \in \mathcal{P}_i$ as follows:

\begin{align}
f^{\text{\it GICP}}({\bm p}'_k, {\bm p}_k, {\bm T}_i, {\bm T}_j) = {\bm d}^T {\bm \Omega} {\bm d}, 
\end{align}

\begin{align}
{\bm d} = {\bm \mu}_k' - {\bm T}_{ij} {\bm \mu}_k, \qquad
{\bm \Omega} &= \left( {\bm C}'_k + {\bm T}_{ij} {\bm C}_k {\bm T}_{ij}^T \right)^{-1},
\end{align}
where ${\bm T}_{ij} = {\bm T}_i^{-1} {\bm T}_j$ is the relative pose between ${\bm T}_i$ and ${\bm T}_j$.

For ease of the following downsampling process, we decompose ${\bm \Omega} = {\bm \Phi} {\bm \Phi}^T$ and use the following function $f^{\text{\it DIST}}$ that gives the same least square errors as $f^{\text{\it GICP}}$:
\begin{align}
f^{\text{\it DIST}}({\bm p}'_k, {\bm p}_k, {\bm T}_i, {\bm T}_j) = {\bm \Phi}^T {\bm d}. 
\end{align}
Because ${\bm \Omega}$ is a symmetric positive definite matrix, the decomposition ${\bm \Omega} = {\bm \Phi} {\bm \Phi}^T$ can efficiently be found via Cholesky decomposition. This modification does not change the final objective function and thus does not affect the optimization process.

\subsection{Fast Coreset Extraction}
\label{sec:coreset}

To extract a weighted subset of input residuals that yields the same quadratic error function as that of the original set, we use an extended version of the efficient coreset set extraction algorithm in \cite{NEURIPS2019_475fbefa}. 

Caratheodory's theorem \cite{cook1972caratheodory} states that every point in a convex hull of points in $\mathbb{R}^D$ can be represented as a weighted sum of a subset of at most $D + 1$ points. Given a Jacobian matrix ${\bm J} = [ {\bm a}_1^T, \cdots, {\bm a}_N^T ]^T$, where ${\bm a}_k = \frac{\partial e_k}{\partial {\bm T}} \in \mathbb{R}^{1 \times D}$ represents derivatives of a residual ${e}_k$, the Hessian matrix ${\bm H} = {\bm J}^T {\bm J}$ can be given by $\sum_{k=1}^{N} {\bm a}_k^T {\bm a}_k$. We consider {\it flattened} vectors ${\bm h}_k \in \mathbb{R}^{D^2}$ of ${\bm a}_k^T {\bm a}_k$ and calculate their mean ${\bm \mu}^h = \frac{\sum_{k=1}^N {\bm h}_k}{N} = \frac{ \sum_{k=1}^N {\text{flatten}(\bm a}_k^T{\bm a}_k)}{N} $. Because ${\bm \mu}^h$ is always in the convex hull of ${\bm h}_k$, we can find a minimum exact coreset (a.k.a., the Caratheodory set) of ${\bm h}_k$ to represent ${\bm \mu}^h$ and scale it to recover the original Hessian matrix from at most $D^2 + 1$ row vectors in the original Jacobian matrix.

\renewcommand{\algorithmicrequire}{\textbf{Input:}}
\renewcommand{\algorithmicensure}{\textbf{Output:}}

\begin{algorithm}[tb]
\small
\caption{Caratheodory(${\bm P}, u, M$)}
\label{alg:caratheodory}
\begin{algorithmic}[1]
\Require
  \Statex A set of points ${\bm P} = \{{\bm p}_1, \cdots, {\bm p}_N\}$ in $\mathbb{R}^{L}$
  \Statex A weighting function $u : {\bm P} \rightarrow [0, \infty]$
  \Statex Target output size $M$
\Ensure
  \Statex A subset of input points and weighting function (${\bm S}, w$) s.t. $\sum \left(u({\bm p}_i) \cdot {\bm p}_i\right) = \sum \left(w({\bm s_i}) \cdot {\bm s_i}\right)$ and $|{\bm S}| = M$
\If{$N \leq M$}
  \State \Return $({\bm P}, u)$
\EndIf
\For{ ${\bm p}_i \in \{{\bm p}_2, \cdots, {\bm p}_N$\} }
  \State ${\bm a}_i \leftarrow {\bm p}_i - {\bm p}_1$
\EndFor
\State ${\bm A} \leftarrow [{\bm a}_2|\cdots|{\bm a}_N]$  \Comment{${\bm A} \in \mathbb{R}^{L \times (N - 1)}$}
\State Compute ${\bm v} = [v_2, \cdots, v_N]^T \neq 0$ s.t. ${\bm A}{\bm v} = 0$
\State $v_1 \leftarrow - \sum_{i=2}^N v_i$
\State $\alpha \leftarrow \min \{ u({\bm p}_i) / v_i | i \in \{1, \cdots, N\} \text{ and } v_i > 0 \}$
\State $w({\bm p}_i) \leftarrow u({\bm p}_i) - \alpha v_i$ for every $i \in \{1, \cdots, N\}$
\State ${\bm S} \leftarrow \{{\bm p}_i | w({\bm p}_i) > 0 \text{ and } i \in \{1, \cdots, N\}\}$
\If{ $|{\bm S} | > M$ }
  \State \Return \Call{Caratheodory}{${\bm S}, w, M$}
\EndIf
\State \Return $({\bm S}, w)$
\end{algorithmic}
\end{algorithm}

\begin{algorithm}[tb]
\small
\caption{Fast-Caratheodory(${\bm P}, u, K, M$)}
\label{alg:fast_caratheodory}
\begin{algorithmic}[1]
\Require
  \Statex A set of points ${\bm P} = \{{\bm p}_1, \cdots, {\bm p}_N\}$ in $\mathbb{R}^{L}$
  \Statex A weighting function $u : {\bm P} \rightarrow [0, \infty]$
  \Statex Number of clusters $K >= L + 2$
  \Statex Target output size $M$
\Ensure
  \Statex A subset of input points and weighting function (${\bm C}, w$) s.t. $\sum \left(u({\bm p}_i) \cdot {\bm p}_i\right) = \sum \left(w({\bm c_i}) \cdot {\bm c_i}\right)$
\If{$| N \leq M |$}
  \State \Return $({\bm P}, u)$
\EndIf
\State $\{ {\bm P}_1, \cdots, {\bm P}_K \} =$ equally divided disjoint subsets of ${\bm P}$
\For{ $i \in \{1, \cdots, K\}$ }
  \State ${\bm \mu}_i \leftarrow \frac{\sum_{{\bm p} \in {\bm P}_i} u({\bm p}) \cdot {\bm p}}{\sum_{{\bm p} \in {\bm P}_i} u({\bm p})}$  \Comment{Compute mean of the cluster}
  \State $u'({\bm \mu_i}) \leftarrow \sum_{{\bm p} \in {\bm P}_i} u({\bm p})$
\EndFor
\State $(\tilde{{\bm \mu}}, \tilde{w}) \leftarrow $ \Call{Caratheodory}{$\{{\bm \mu}_1, \cdots, {\bm \mu}_K\}, u'$}
\State ${\bm C} \leftarrow \cup_{{\bm \mu}_i \in \tilde{\bm \mu}} {\bm P}_i$   \Comment{Recover points from selected clusters}
\For{${\bm \mu}_i \in \tilde{\bm \mu}$ and ${\bm p} \in {\bm P}_i$}
  \State $w({\bm p}) \leftarrow \frac{\tilde{w}({\bm \mu}_i) u({\bm p})}{\sum_{{\bm p} \in {\bm P}_i} u({\bm p})}$
\EndFor
\State \Return \Call{Fast-Caratheodory}{${\bm C}, w, K, M$} 
\end{algorithmic}
\end{algorithm}

\begin{algorithm}[tb]
\small
\caption{Fast-Caratheodory-Quadratic(${\bm e}, {\bm J}, M$)}
\label{alg:fast_caratheodory_quadratic}
\begin{algorithmic}[1]
\Require
  \Statex A residual vector ${\bm e} \in \mathbb{R}^N = [e_1, \cdots, e_N]^T$
  \Statex Jacobian matrix ${\bm J} \in \mathbb{R}^{N \times D} = [{\bm a}_1, \cdots, {\bm a}_N]^T$
  \Statex Target output size $M$
\Ensure
  \Statex A function $g$ to select a subset of ${\bm e}$
  \Statex Weights ${\bm w}$ for the selected subset
  \Statex $g$ and ${\bm w}$ satisfy $\tilde{\bm J}^T {\bm W} \tilde{\bm J} = {\bm J}^T {\bm J}$, $\tilde{\bm J}^T {\bm W} \tilde{\bm e} = {\bm J}^T {\bm e}$, $\tilde{\bm e}^T {\bm W} \tilde{\bm e} = {\bm e}^T {\bm e}$, where $\tilde{\bm e} = g({\bm e}), \tilde{\bm J} = g({\bm J})$, and ${\bm W} = \text{diag}({\bm w})$
\For{ $i \in \{ 1, \cdots, N \}$ }
  \State ${\bm h}_i \leftarrow $ upper triangular elements of ${\bm a}_i^T {\bm a}_i$
  \State ${\bm b}_i \leftarrow {\bm a}_i^T e_i$
  \State $c_i \leftarrow e_i^2$
  \State ${\bm p}_i \leftarrow [{\bm h}_i^T, {\bm b}_i^T, c_i]^T$
  \State $u({\bm p}_i) \leftarrow 1 / N$
\EndFor
\State ${\bm P} \leftarrow \{{\bm p}_1, \cdots, {\bm p}_N\}$
\State $({\bm C}, w) \leftarrow$ \Call{Fast-Caratheodory}{${\bm P}, u, K, M$}
\State $g \leftarrow $ A function to select entries of ${\bm e}$ exist in ${\bm C}$
\State ${\bm w} \leftarrow [ w(c_i) | c_i \in g({\bm e}) ]$
\State \Return $(g, {\bm w})$
\end{algorithmic}
\end{algorithm}

Algorithms \ref{alg:caratheodory}, \ref{alg:fast_caratheodory}, and \ref{alg:fast_caratheodory_quadratic} describe the proposed coreset extraction algorithm extending the algorithms in \cite{cook1972caratheodory,NEURIPS2019_475fbefa}. Due to space limitations, we provide only a brief explanation of the algorithms and highlight the modifications we made. For details of the original algorithms, we refer the reader to \cite{NEURIPS2019_475fbefa}. The code for the proposed algorithm is available online so that the reader can confirm the details \footnote{\url{https://github.com/koide3/caratheodory2}}.

Algorithm \ref{alg:caratheodory} is Caratheodory's classic algorithm for finding a minimum exact coreset from a weighted input set \cite{cook1972caratheodory}. This algorithm finds and eliminates redundant data points one by one. Because the time required to execute the algorithm is $O(N^2 L^2)$, it cannot be directly applied to large datasets. To reduce the time complexity, Algorithm \ref{alg:fast_caratheodory} divides the input set into $K$ disjoint subsets (e.g., $K=64$) and computes the means of the clusters. It then applies Algorithm \ref{alg:caratheodory} to extract a Caratheodory set of the cluster means and eliminates clusters that are not selected to be in the Caratheodory set. It recovers the set of input data from the clusters in the extracted Caratheodory set and repeats this process until the number of data points becomes less than or equal to a target output size. The clustering strategy significantly improves the time complexity and can find a Caratheodory set in time linear to the number of input data points $O(NL)$. Algorithm \ref{alg:fast_caratheodory_quadratic} takes as input a residual vector ${\bm e}$ and its Jacobian matrix ${\bm J}$, computes {\it flattened} vectors of ${\bm H}, {\bm b}, c$, and passes them to Algorithm \ref{alg:fast_caratheodory} to obtain a weighted subset of input residuals to reconstruct the original quadratic function.

The modifications we made to fit the algorithms in \cite{NEURIPS2019_475fbefa} to our problem can be summarized as follows:
\begin{enumerate}
  \item While the original algorithm selects only a minimum number of data points to compose a Caratheodory set, we modified Algorithms \ref{alg:caratheodory} and \ref{alg:fast_caratheodory} so that they can change the target number of data points $M$ to control the trade-off between approximation accuracy and computation speed. By increasing $M$, the algorithm extracts more data that are redundant to compose a Caratheodory set but help approximate the nonlinearity of the original registration error function. Note that the proposed algorithm does not extract exactly $M$ points but rather extracts points in the range [$\max(M - K, 29), M$]. While we can easily modify Algorithm \ref{alg:fast_caratheodory} to select exactly $M$ points, we avoided doing so to save processing time.
  \item In the original implementation of Algorithm \ref{alg:caratheodory}, most of the computation time was taken by SVD to find $v \neq 0$ such that ${\bm A}v = 0$. To improve the processing speed, we use an efficient LU decomposition method to find the nullspace of ${\bm A}$ instead of SVD.
  \item While \cite{NEURIPS2019_475fbefa} recovers only the Hessian matrix ${\bm H}$, we recover ${\bm b}$ and $c$ in addition to ${\bm H}$ to approximate the quadratic function in Eq. \ref{eq:quadratic}. Although this modification increases the dimension of the input vectors from $D^2$ to $D^2 + D + 1$, we can omit the non-upper-triangular elements of ${\bm H}$, since it is symmetric. As a result, for a function with six-dimensional input, the input vector dimension becomes $21 + 6 + 1 = 28$, and we need only $28 + 1 = 29$ residuals to exactly recover the original quadratic function at a minimum.
\end{enumerate}
As we will show in Sec. \ref{sec:validation}, modifications 2 and 3 reduce processing time by a factor of 10. Note that because the registration error function is symmetric for ${\bm T}_i$ and ${\bm T}_j$, the subset extracted for ${\bm T}_j$ exactly reconstructs ${\bm H}_{ii}, {\bm H}_{ij}$, and ${\bm b}_{i}$ as well.

\subsection{Point Cloud Downsampling}

\begin{algorithm}[tb]
\small
\caption{Exact-Downsampling($\mathcal{P}_i, \mathcal{P}_j, \breve{\bm T}_i, \breve{\bm T}_j, f^{\text{\it REG}}, M$)}
\label{alg:downsampling}
\begin{algorithmic}[1]
\Require
  \Statex Point clouds $\mathcal{P}_i$ and $\mathcal{P}_j$
  \Statex Poses $\breve{\bm T}_i$ and $\breve{\bm T}_j$ to evaluate errors between $\mathcal{P}_i$ and $\mathcal{P}_j$
  \Statex Registration error function $f^{\text{\it REG}}$
  \Statex Target output size $M$
\Ensure
  \Statex A function $g$ to select a subset of residuals of input points
  \Statex A weight vector ${\bm w}$
\State Shuffle points in $\mathcal{P}_j$
\State ${\bm e}^T {\bm e} \leftarrow f^{\text{\it REG}}(\mathcal{P}_i, \mathcal{P}_j, {\bm x}_i, {\bm x}_j)$
\State ${\bm J} \leftarrow \partial {\bm e} / \partial {\bm T}_j$ 
\State $(g, {\bm w}) \leftarrow \Call{Fast-Caratheodory-Quadratic}{{\bm e}, {\bm J}, M}$
\State \Return $(g, {\bm w})$
\end{algorithmic}
\end{algorithm}

\begin{figure}[tb]
  \centering
  \begin{minipage}[tb]{0.49\linewidth}
  \centering
  \includegraphics[width=\linewidth]{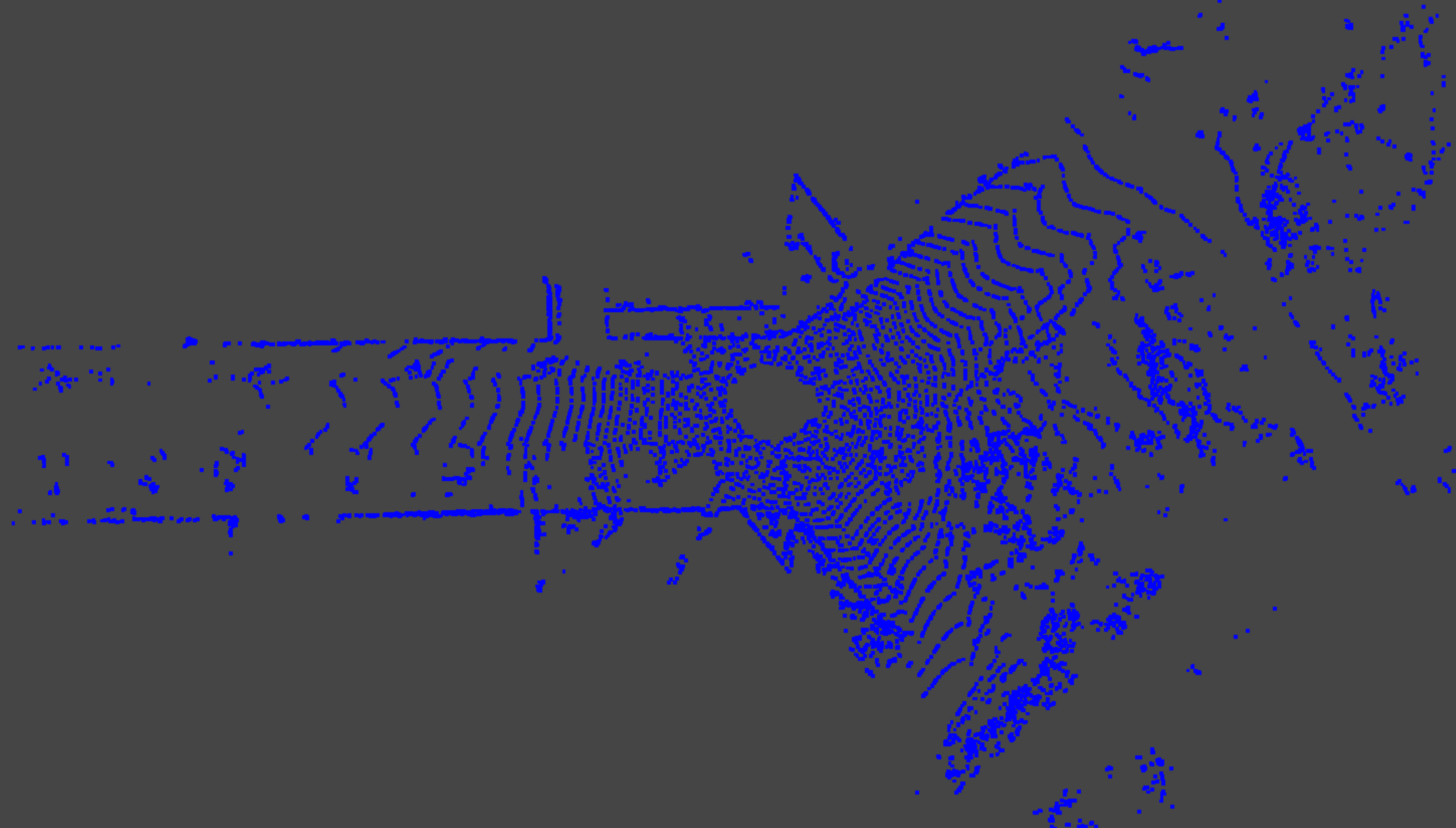}
  \subcaption{Target points}
  \end{minipage}
  \begin{minipage}[tb]{0.49\linewidth}
  \centering
  \includegraphics[width=\linewidth]{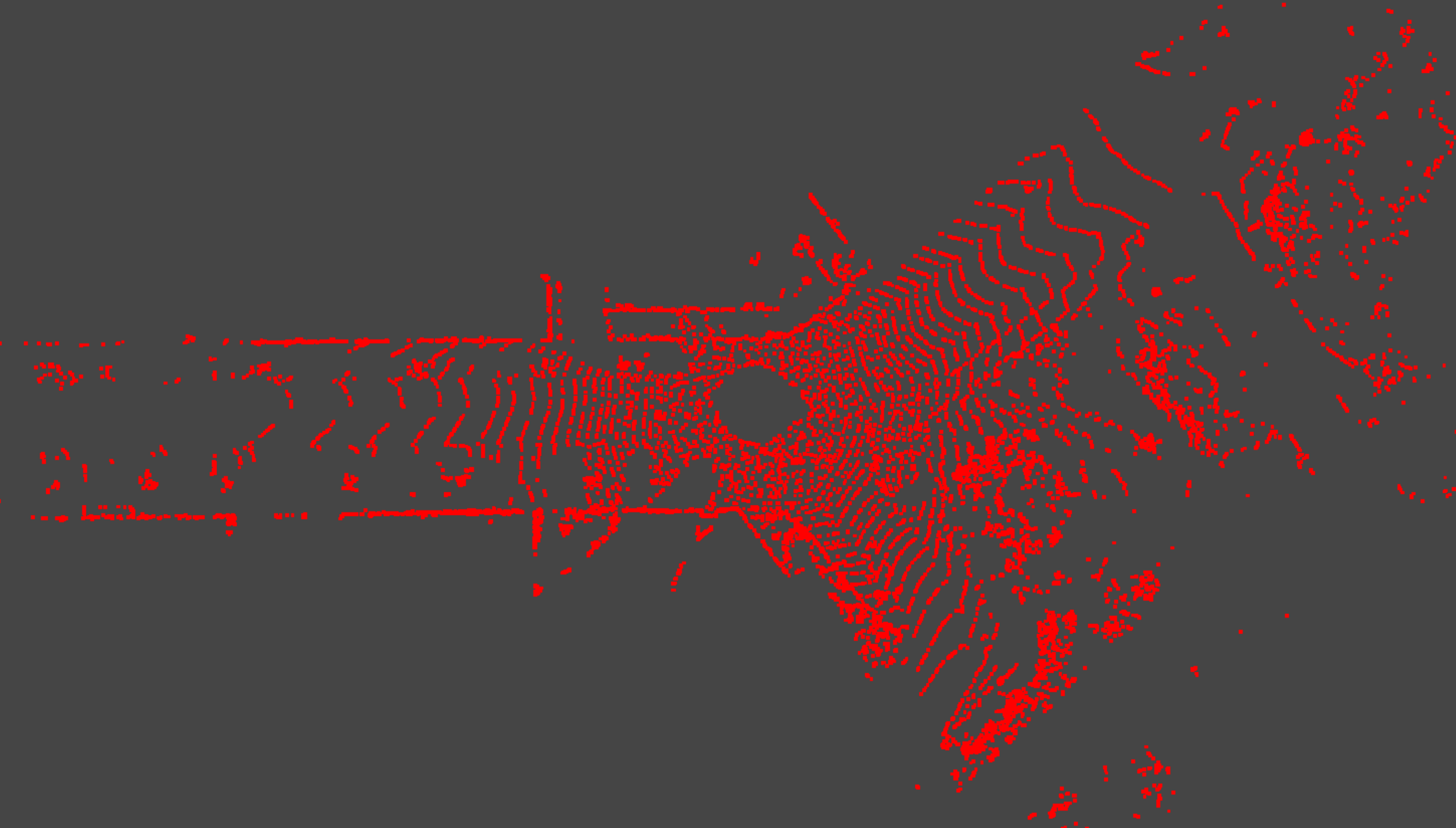}
  \subcaption{Source points (N = 10,000)}
  \end{minipage}
  \begin{minipage}[tb]{0.49\linewidth}
  \centering
  \includegraphics[width=\linewidth]{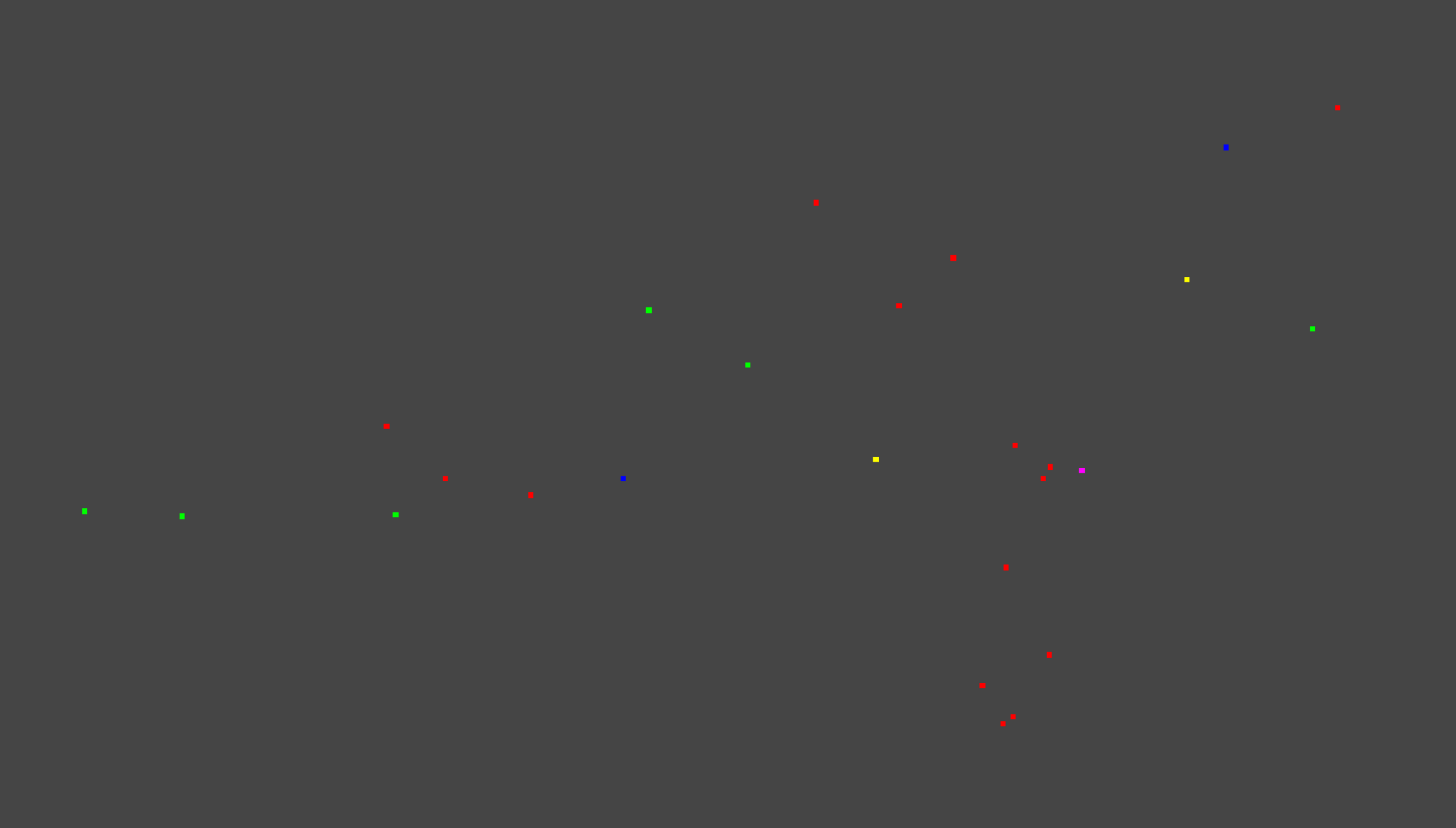}
  \subcaption{Sampled residuals (M = 29)}
  \end{minipage}
  \begin{minipage}[tb]{0.49\linewidth}
  \centering
  \includegraphics[width=\linewidth]{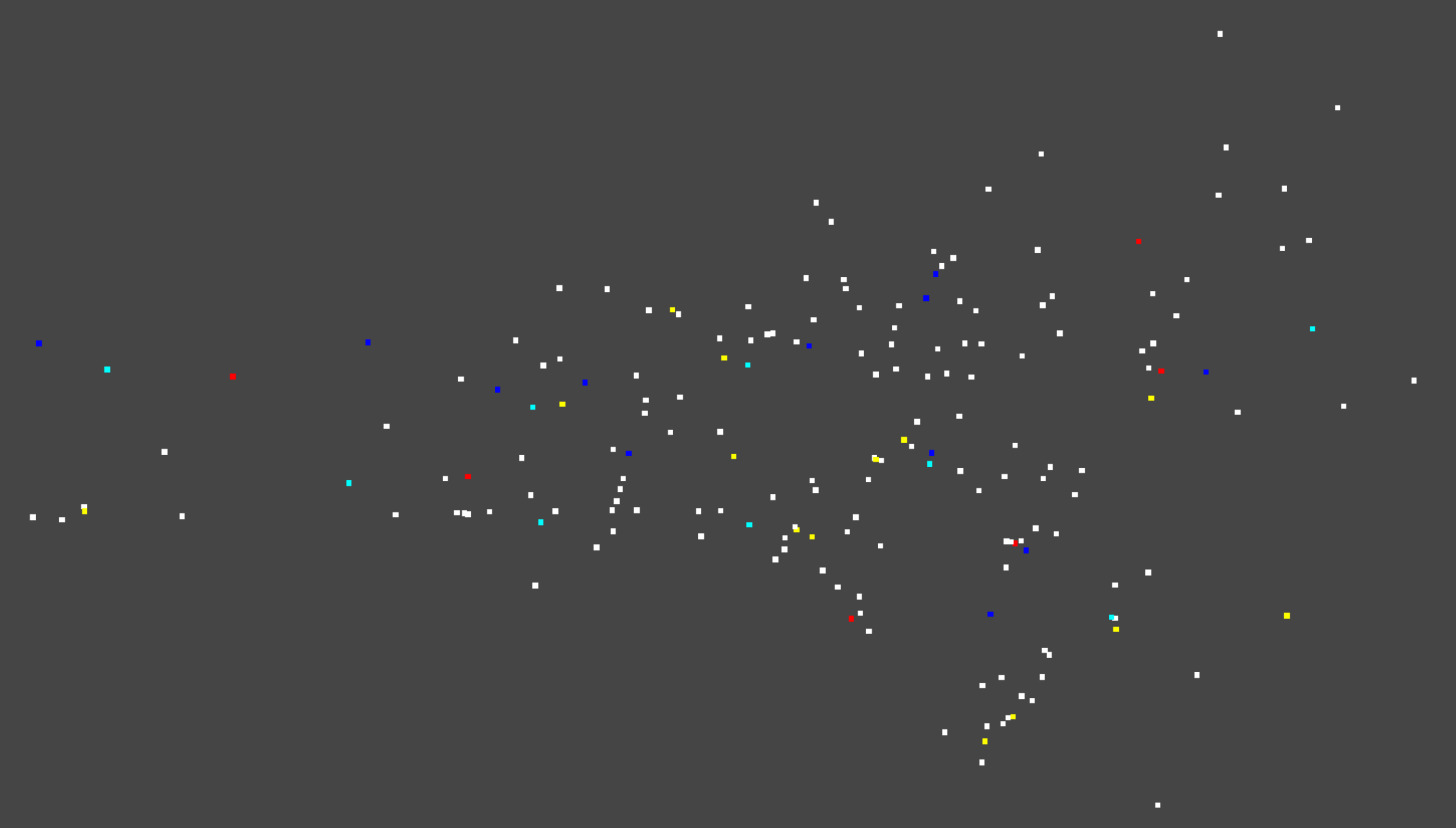}
  \subcaption{Sampled residuals (M = 512)}
  \end{minipage}
  \caption{Example of downsampling results. Source points (b) and sampled points with different target numbers of residuals (M=29 (c), and M=512 (d)) all yield the same quadratic registration error function for the target points at the evaluation point. The colors of the sampled points represent the selected axes of residuals (R=X, G=Y, B=Z).}
  \label{fig:samples}
\end{figure}

The proposed exact point cloud downsampling algorithm is summarized as in Algorithm \ref{alg:downsampling}. Because Algorithm \ref{alg:fast_caratheodory} reduces the dataset size by dropping contiguous items, we first shuffle the input points to avoid bias in the sampling result. We then evaluate the residuals ${\bm e}$ and Jacobian ${\bm J}$ of the registration error function at $\breve {\bm T}_i$ and ${\breve{\bm T}_j}$. We input ${\bm e}$ and ${\bm J}$ to Algorithm \ref{alg:fast_caratheodory_quadratic} and obtain a subset of the residuals and weights.

During Gauss-Newton optimization, we iteratively re-linearize the error function $f^{\text{\it REG}}$ with the extracted subset to obtain a quadratic error function.

Fig. \ref{fig:samples} (c) and (d) show examples of downsampling results. The colors of the points indicate the selected axes of the sampled points (R=X, G=Y, B=Z). Because the proposed algorithm works on a per-residual basis, it happens that only one or two axes of a point are selected (Fig. \ref{fig:samples} (c)). Because the algorithm tends to drop contiguous residuals, most of the selected points have all selected axes when the number of target residuals is sufficiently large (Fig. \ref{fig:samples} (d)). Both the sampled points with $M=29$ and $M=512$ shown in Fig. \ref{fig:samples} (c) and (d) exactly recover the quadratic registration error function between the original input point clouds shown in Fig. \ref{fig:samples} (a) and (b).

\section{Experiments}

\subsection{Numerical Validation}
\label{sec:validation}





%
%
%
%
%
%

We first verified the proposed algorithm through numerical validation. We began by randomly generating ${\bm J} \in \mathbb{R}^{N \times 6}$ and ${\bm e} \in \mathbb{R}^{N}$ ($N = 30{,}000 \approx 10{,}000$ points) and applying the proposed downsampling algorithm to obtain a weighted subset of the residuals $(g, {\bm w})$. We then calculated the quadratic function approximation error as $\max(\|{\bm J}^T {\bm J} - \tilde{\bm J}^T {\bm W} \tilde{\bm J}\|, \|{\bm J}^T {\bm e} - \tilde{\bm J}^T {\bm W} \tilde{\bm e}\|, {\bm e}^T{\bm e} - \tilde{\bm e}^T {\bm W} \tilde{\bm e})$, where $\tilde{\bm J} = g({\bm J}), \tilde{\bm e} = g({\bm e})$, and ${\bm W = \text{diag}({\bm w})}$. We repeated this numerical validation 100 times \footnote{The code for reproducing this experiment is available at the GitHub repository.}.

To evaluate the speed gain of the modifications proposed in Sec. \ref{sec:coreset}, we ran the algorithm with three configurations:
\begin{enumerate}
  \item Naive implementation of \cite{NEURIPS2019_475fbefa} with the expanded input dimension $D^2 + D + 1 = 43$
  \item Naive implementation of \cite{NEURIPS2019_475fbefa} with the compact input dimension without the non-upper-triangular elements of the Hessian ($(D + 1) \times (D / 2) + D + 1 = 28$)
  \item Using the compact input dimension (28) and LU decomposition instead of SVD
\end{enumerate}

All three configurations produced approximation errors below $10^{-10}$ for all trials, showing the validity of the proposed algorithm. Configuration 1 took approximately 125 ms for each trial. With configuration 2, the processing time decreased to 76 ms by using the compact input representation. Finally, with configuration 3, the processing time further decreased to 7 ms, or roughly 5.6\% of the time for configuration 1.

We then evaluated the change in processing time for the proposed algorithm when the target output size varies from 29 to 1024. Again, all the selected subsets showed reconstruction errors below $10^{-10}$ for all the settings. Table \ref{tab:time_analysis} summarizes the average processing times with several target numbers of residuals $M$. As shown, when the target number of residuals was increased, the processing time also grew, albeit gradually. While all the settings exactly reconstructed the original quadratic error function at the evaluation point, as we will show in the following Sec. \ref{sec:factor_analysis}, a larger number of residuals leads to better approximation accuracy under pose displacements. Thus, setting the target number of residuals parameter involves a trade-off between approximation accuracy and processing speed.

\begin{table}[tb]
  \caption{Processing time for the exact sampling algorithm}
  \label{tab:time_analysis}
  \centering
  \scriptsize
  \begin{tabular}{l|lllllll}
  \toprule
  Number of residuals (M) & 29 & 64 & 128 & 256 & 512 & 1024 \\
  Processing time [ms] & 6.87 & 13.39 & 19.34 & 25.03 & 29.99 & 34.19 \\
  \bottomrule
  \end{tabular}
\end{table}

\subsection{Nonlinearity Approximation Accuracy Analysis}
\label{sec:factor_analysis}

We next compared the approximation accuracy of the proposed exact sampling with that of the commonly used random sampling approach. Here, we took two consecutive frames in the sequence 00 of the KITTI dataset \cite{geiger2013vision}, as shown in Fig. \ref{fig:samples}.
We aligned the frames by using standard GICP scan matching and then sampled subsets of the input points and residuals using random sampling and the proposed exact sampling method while changing the target output size.

We computed the Hessian matrices of $f^{\text{\it REG}}$ by using the sampled subsets and compared them with that computed using the original input points. To compare Hessian matrices, we used the normalized KLD metric defined as follows:
\begin{align}
\text{KLD}({\bm H}, \tilde{\bm H}) = \frac{1}{2} \left( \log(\frac{|{\bm H}|}{|\tilde{\bm H}|}) + \text{tr}( {\bm H}^{-1} \tilde{\bm H} ) \right), \\
\text{NormedKLD}({\bm H}, \tilde{\bm H}) = 1 - \exp(-\text{KLD}({\bm H}, \tilde{\bm H})).
\end{align}
We repeated the evaluation 100 times while shuffling the order of the points.

\begin{table}[tb]
  \caption{Hessian approximation accuracy evaluation result}
  \label{tab:kld_analysis}
  \centering
  \scriptsize
  \begin{tabular}{l|ll}
  \toprule
  Method          & Num. of points / residuals       & Normalized KLD  \\
  \midrule
  \midrule
                  & 10 points $\approx$ 30 residuals     & 0.996 $\pm$ 0.026 \\
                  & 64 points $\approx$ 192 residuals    & 0.437 $\pm$ 0.213 \\
  Random sampling & 256 points $\approx$ 768 residuals   & 0.107 $\pm$ 0.062 \\
                  & 1024 points $\approx$ 3072 residuals & 0.024 $\pm$ 0.013 \\
                  & 10149 points (original points)       & 0.000 $\pm$ 0.000 \\
  \midrule
  Exact sampling  & 29 - 3072 residuals                  & 0.000 $\pm$ 0.000 \\
  \bottomrule
  \end{tabular}
\end{table}

Table \ref{tab:kld_analysis} summarizes the normalized KLDs for the random and exact sampling results. The random sampling results showed large errors when the number of points was small. In particular, when the number of points was set to 10, we observed that the Hessian matrix computed from the sampled points sometimes degenerated. As the number of points increased, the approximation accuracy of the random sampling results improved. When and only when the target number of points was set to the same number as the number of input points, the approximation error became zero. This result suggests that random sampling would not be a good choice for sampling point cloud registration errors, since it substantially changes the objective function shape when the number of sampling points is small. On the other hand, as verified in Sec. \ref{sec:validation}, the proposed algorithm exactly recovered the original Hessian matrix with a small number of residuals and thus the normalized KLD was always zero at the evaluation point.

We then evaluated the changes in function approximation accuracy of the two sampling methods under pose displacements. We applied random rotation noise and re-evaluated the registration error function using the sampled subsets to obtain ${\bm H}$ and ${\bm b}$. We then computed the displacement vector claimed by the linearized system (${\Delta {\bm x}} = {\bm H}^{-1}{\bm b}$) and compared it with that computed using the original points.

\begin{figure}[tb]
  \centering
  \begin{minipage}[tb]{\linewidth}
  \centering
  \includegraphics[width=\linewidth]{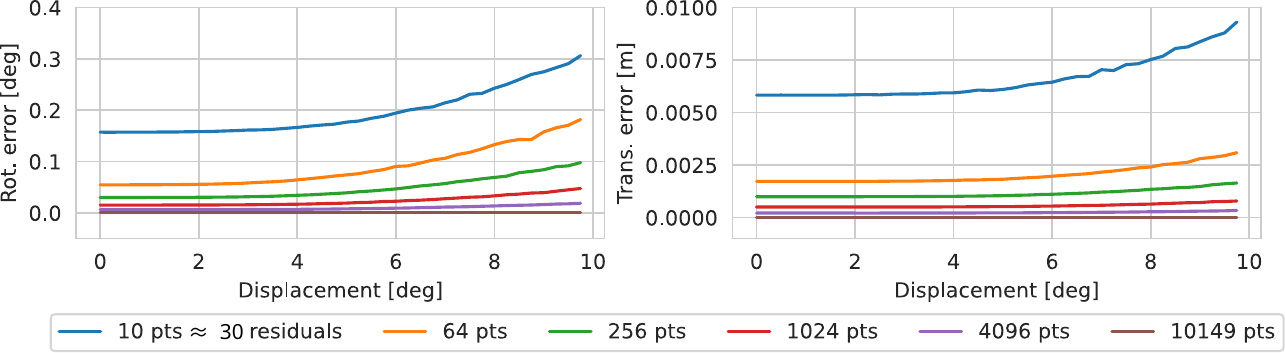}
  \subcaption{Random sampling}
  \end{minipage}
  \begin{minipage}[tb]{\linewidth}
  \centering
  \includegraphics[width=\linewidth]{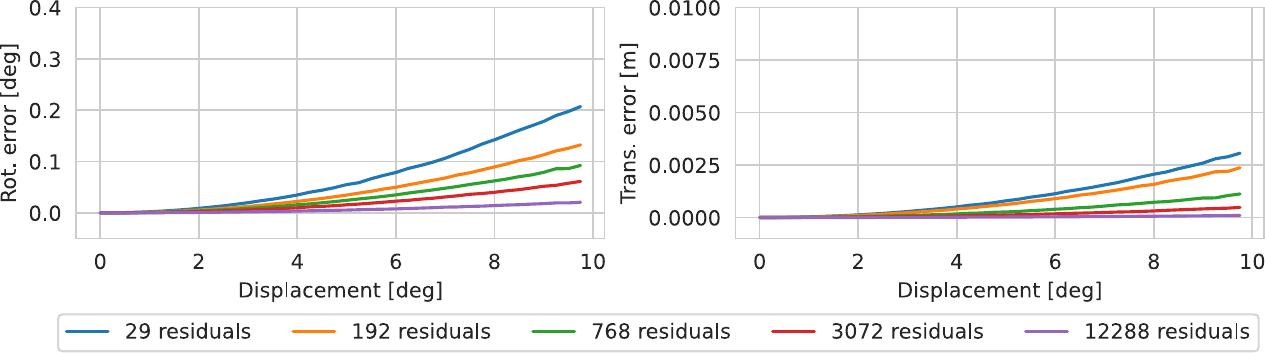}
  \subcaption{Exact sampling}
  \end{minipage}
  \caption{Errors of the displacement vector $\Delta {\bm x} = {\bm H}^{-1}{\bm b}$ of random and exact sampling results under rotation noise.}
  \label{fig:errors}
\end{figure}

Fig. \ref{fig:errors} summarizes the displacement vector errors. Because the residuals sampled by the proposed algorithm exactly recovered the original quadratic error function, the proposed algorithm showed zero displacement vector errors with all the settings when the noise was zero. As the rotation noise became larger and the current estimate separated further from the evaluation point, the displacement vector errors became gradually worse due to the nonlinearity of the registration error function. However, we can see that the results of the proposed exact sampling show substantially smaller errors compared to the random sampling results with equivalent output size settings.

\begin{table*}[tb]
  \caption{ATEs [m] of pose graph optimization and global registration error minimization results}
  \label{tab:apes}
  \centering
  \scriptsize
  \begin{tabular}{l|llll}
  \toprule
  \multirow{2}{*}{Method} & \multicolumn{4}{c}{Sequence} \\
                          & 00                  & 02                   & 05                  & 08                  \\
  \midrule
  \midrule
  Pose graph optimization (Identity)        & 1.6257 $\pm$ 0.7529 & 23.9856 $\pm$ 8.2752 & 1.5149 $\pm$ 0.6760 & 9.3636 $\pm$ 3.2890 \\
  Pose graph optimization (Hessian)         & 1.3777 $\pm$ 0.6051 & 9.3406  $\pm$ 3.1490 & 1.6249 $\pm$ 0.8489 & 5.0532 $\pm$ 2.5991 \\
  Pose graph optimization (Hessian + Dense) & 1.1846 $\pm$ 0.5625 & 9.3393  $\pm$ 3.1486 & 1.6240 $\pm$ 0.8488 & 5.0520 $\pm$ 2.5993 \\
  \midrule
  Registration error minimization + Exact sampling (M=29)                     & 0.9553 $\pm$ 0.4650 & 8.9679  $\pm$ 3.0856 & 0.2917 $\pm$ 0.1060 & 4.4394 $\pm$ 2.5294 \\
  \bottomrule
  \end{tabular}
\end{table*}

\subsection{Application to Global Trajectory Optimization}

We applied the proposed exact downsampling to global trajectory optimization on the longest four sequences (00, 02, 05, 08) in the KITTI dataset \cite{geiger2013vision} in order to compare its trajectory estimation accuracy and associated processing time with those of the conventional pose graph optimization.

{\bf Evaluation protocol}: To obtain an initial guess of sensor poses, we applied GICP scan matching between consecutive frames (i.e., odometry estimation). Meanwhile, we ran ScanContext \cite{Kim_2018} to detect loop candidates and validated them via GICP scan matching. We constructed a pose graph with relative pose constraints between the consecutive frames and loop pairs and performed optimization using the Levenberg-Marquardt optimizer in GTSAM \cite{gtsam}. The objective function of pose graph optimization is defined as follows:
\begin{align}
f^{\text{\it PGO}}(\mathcal{X}) &= \sum_{({\bm T}_i, {\bm T}_j, \hat{\bm T}_{ij})} \| f^{\text{\it RP}}({\bm T}_i, {\bm T}_j, \hat{\bm T}_{ij}) \|^2, \\
f^{\text{\it RP}}({\bm T}_i, {\bm T}_j, \hat{\bm T}_{ij}) &= \rho \left( {\bm d}_{ij}^T {\bm H}_{ij} {\bm d}_{ij} \right), \\
{\bm d}_{ij} &= \log \left( \hat{\bm T}_{ij}^{-1} {\bm T}_i^{-1} {\bm T}_j \right),
\end{align}
where $\mathcal{X}$ is the set of sensor poses, $\hat{\bm T}_{ij}$ is the relative pose measurement given by scan matching, and $\rho$ is Cauchy's robust kernel. For ${\bm H}_{ij}$, we used two settings: 1) setting the identity matrix ${\bm I}_{6 \times 6}$ to ${\bm H}_{ij}$, and 2) setting the Hessian matrix at the last iteration of GICP scan matching to ${\bm H}_{ij}$.

After performing pose graph optimization, we found all overlapping point cloud pairs by voxel-based overlap assessment \cite{Koide2021} and constructed a densely connected graph with factors between all detected overlapping pairs, as shown in Fig. \ref{fig:graph}. With this dense graph structure, we again optimized the sensor poses using pose graph optimization and global registration error minimization. The objective function for global registration error minimization is defined as

\begin{align}
f^{\text{\it GR}}(\mathcal{X}) &= \sum_{({\bm T}_i, {\bm T}_j) \in \mathcal{X}^O} \| f^{\text{\it REG}}(\mathcal{P}_i, \mathcal{P}_j, {\bm T}_i, {\bm T}_j) \|^2,
\end{align}
where $\mathcal{X}^O$ is the set of point cloud pairs with an overlap.

For pose graph optimization, we applied GICP scan matching for all frame pairs and created relative pose constraints from the scan matching results.
For global registration error minimization, we applied the exact sampling for all frame pairs with the target number of residuals $M=29$ and optimized the sensor poses by minimizing the global registration errors computed with the sampled residuals. For the sequence 00, we also ran the same algorithm with M=256 and M=1024 and the original input points to determine the effect that changing the number of samples has on processing time and estimation accuracy.

{\bf Comparison with pose graph optimization}: Table \ref{tab:apes} summarizes the absolute trajectory errors (ATEs) \cite{Zhang_2018} for pose graph optimization and global registration error minimization. We used the {\it evo} toolkit \footnote{\url{https://github.com/MichaelGrupp/evo}} to measure the ATEs.

For most of the sequences, pose graph optimization with identity Hessian matrices produced much worse results than when Hessian matrices composed of the scan matching results were used. This suggests the importance of appropriately modeling the relative pose constraints for better estimation results.

\begin{figure}[tb]
  \centering
  \includegraphics[width=0.6\linewidth]{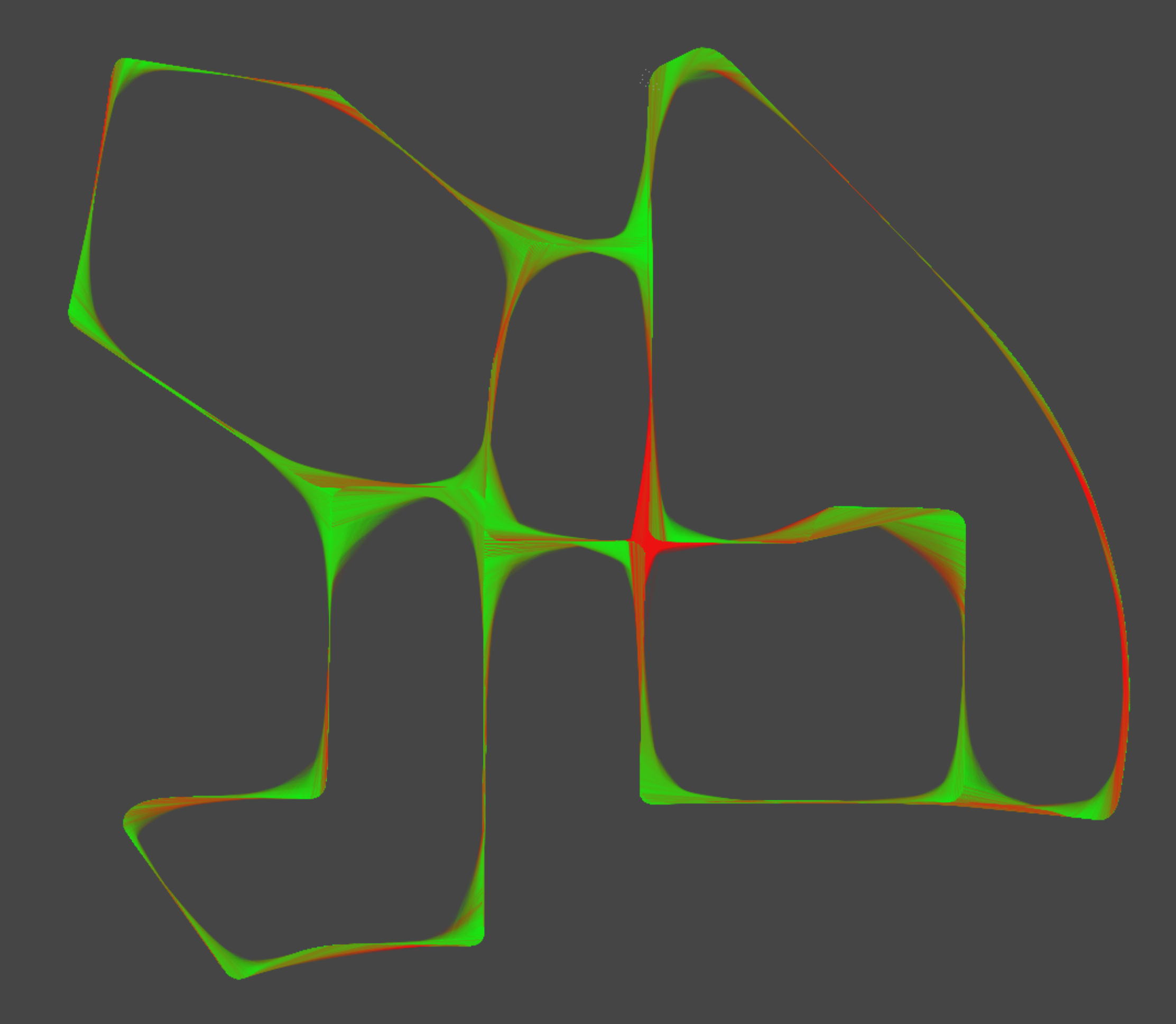}
  \caption{Dense pose graph optimization result. Color indicates the magnitude of the relative pose errors of each factor (Green: small error, Red: large error).}
  \label{fig:deteriorated_graph}
\end{figure}

Although pose graph optimization on the dense factor graph showed better ATEs than those for the sparse graph, the accuracy gain was not very significant. Fig. \ref{fig:deteriorated_graph} shows the dense pose graph optimization graph. The colors of the factors indicate the magnitudes of the relative pose errors. We can see that the factors between frames at a distance tend to show large errors because the GICP scan matching failed on small overlapping frames. This suggests that increasing the number of factors does not always improve the estimation accuracy of pose graph optimization.

The global registration error minimization approach greatly improved the ATEs of all the sequences compared to those of pose graph optimization. Because this approach directly minimizes registration errors over the entire map while avoiding the Gaussian approximation of the relative pose constraint, it can accurately constrain the relative pose between frames with small overlap.

\begin{table}[tb]
  \caption{ATEs of global registration error minimization with different sampling settings for the sequence 00}
  \label{tab:apes00}
  \centering
  \scriptsize
  \begin{tabular}{l|c|c}
  \toprule
  Method & ATE [m]                & Optimization time [h] \\
  \midrule
  \midrule
  Exact sampling (M=29)              & 0.9553 $\pm$ 0.4650 & 1.67 \\
  Exact sampling (M=256)             & 0.9549 $\pm$ 0.4646 & 1.88 \\
  Exact sampling (M=1024)            & 0.9549 $\pm$ 0.4647 & 2.36 \\
  \midrule
  Original points (10,000 points) & 0.9549 $\pm$ 0.4647 & 13.21 \\
  \bottomrule
  \end{tabular}
\end{table}

{\bf Effect of the exact sampling}: Table \ref{tab:apes00} summarizes the ATEs of global registration error minimization with different sampling settings. While the ATEs improved as the number of residuals increased, the smallest number of residuals M=29 showed an ATE very close to that of the original points. Because we started the optimization from a sufficiently good initial guess, the approximation errors were sufficiently small for the small displacements with the minimum number of residuals (M=29) as shown in Sec. \ref{sec:factor_analysis}. We consider that, if the initial guess had larger errors, the results from the smallest sampling setting might deteriorate and the other sampling settings with more residual samples would show better results.

\begin{table}[tb]
  \caption{Memory consumption}
  \label{tab:memory_analysis}
  \centering
  \scriptsize
  \begin{tabular}{l|l|ll}
  \toprule
  \multirow{2}{*}{Sampling method} & \multirow{2}{*}{Num. residuals} & \multicolumn{2}{c}{Memory consumption [GB]} \\
                                   &                                 & Correspondences & Sampled residuals \\
  \midrule
  \midrule
  \multirow{3}{*}{Exact sampling}
                   & 29              & 0.11            & 0.14 \\
                   & 256             & 0.43            & 1.15 \\
                   & 1024            & 1.52            & 4.87 \\
  \midrule
  All residuals    & $\sim$30,000   & 25.13           & 0.00 \\
  \bottomrule
  \end{tabular}
\end{table}

\begin{figure}[tb]
  \centering
  \includegraphics[width=\linewidth]{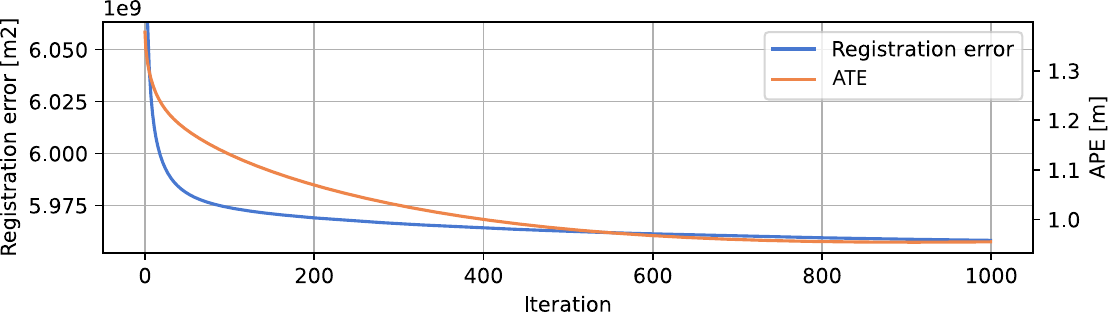}
  \caption{Global registration error and ATE change during global registration error minimization.}
  \label{fig:optimization_curve}
\end{figure}

{\bf Memory consumption}: Table \ref{tab:memory_analysis} summarizes the memory consumption of global registration error minimization for each sampling setting. For the case in which the original points are used, where we need to remember the point correspondences for each factor, approximately 25 GB of memory was required. With exact sampling, where, in addition to remembering the point correspondences, we need to remember the selected residual samples, total memory consumption was substantially smaller. For the setting M=29, only 0.11 GB was needed for the point correspondences and only 0.14 GB was needed for the sampled residuals.

\begin{figure}[tb]
  \centering
  \includegraphics[width=\linewidth]{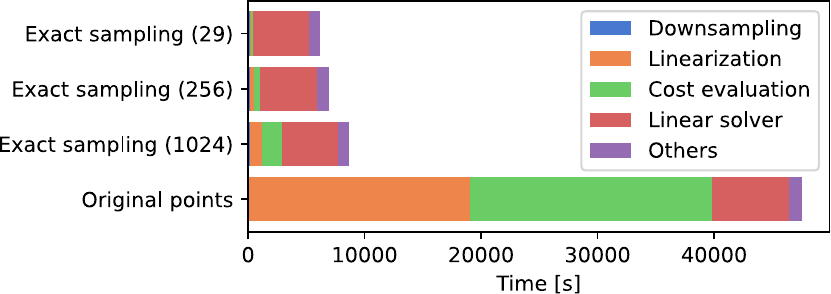}
  \caption{Optimization time breakdown.}
  \label{fig:optimization_time_analysis}
\end{figure}

{\bf Processing time}: Fig. \ref{fig:optimization_curve} shows the decreasing pattern of both the global registration error and the ATE during optimization. With this large-scale factor graph optimization, more than several hundred iterations were needed before convergence. Fig. \ref{fig:optimization_time_analysis} shows a breakdown of the optimization times for the different sampling settings. With the original points, the optimization took approximately 13.2 hours on a CPU with 128 threads. We can see that the linearization and cost evaluation of the registration error function consumed a majority of the optimization time. Although the proposed exact sampling introduced additional processing time for downsampling of approximately 130 s, it drastically decreased the processing time for linearization and cost evaluation and reduced the total optimization time (1.67 hours with M=29, 2.36 hours with M=1024). Note that while we used the multi-threaded MULTIFRONTAL\_CHOLESKY linear solver in GTSAM, we observed that it did not fully utilize all the available CPU cores. We consider that the total optimization time can be further improved by using a linear solver dedicated to multi-threading or GPU computing.

\section{Conclusion}

This work proposed a point cloud downsampling algorithm based on an algorithm for efficient exact coreset extraction. Experimental results showed that this approach drastically reduces the linearization cost of the point cloud registration error function without sacrificing accuracy. We plan to apply the proposed algorithm to extremely large city- or nation-scale mapping problems and other types of data (e.g., dense full BA for visual SLAM).

\bibliographystyle{IEEEtran}
\bibliography{iros2023}

\end{document}